\documentclass[letterpaper, 10 pt, conference]{ieeeconf}  

\IEEEoverridecommandlockouts                              

\usepackage{cite}
\usepackage{amsmath,amssymb,amsfonts}

\usepackage{graphicx}
\usepackage{textcomp}
\usepackage{xcolor}
\usepackage{algorithmic}
\usepackage[linesnumbered,ruled,vlined]{algorithm2e}
\usepackage{booktabs}
\usepackage{array}
\usepackage{booktabs}
\usepackage{footnote}
\usepackage{tabularx}
\usepackage{threeparttable}
\usepackage{ragged2e}
\makesavenoteenv{tabular}
\usepackage{calc}
\def\BibTeX{{\rm B\kern-.05em{\sc i\kern-.025em b}\kern-.08em
    T\kern-.1667em\lower.7ex\hbox{E}\kern-.125emX}}

\begin{document}
\onecolumn
\title{A Large Language Model-Enhanced Q-learning for Capacitated Vehicle Routing Problem with Time Windows\\
}


\author{Linjiang Cao, Maonan Wang and Xi Xiong
\thanks{This work was supported in part by NSFC Project 72371172 and Fundamental Research Funds for the Central Universities.}
\thanks{L. Cao and X. Xiong are with the Key Laboratory of Road and Traffic Engineering, Ministry of Education, Tongji University, Shanghai, China. M. Wang is with the School of Science and Engineering, the Chinese University of Hong Kong, Shenzhen, China (Emails: 2431743@tongji.edu.cn, maonanwang@link.cuhk.edu.cn, xi\_xiong@tongji.edu.cn,)}
}

\maketitle

\begin{abstract}
The Capacitated Vehicle Routing Problem with Time Windows (CVRPTW) is a classic NP-hard combinatorial optimization problem widely applied in logistics distribution and transportation management. 
Its complexity stems from the constraints of vehicle capacity and time windows, which pose significant challenges to traditional approaches. 
Advances in Large Language Models (LLMs) provide new possibilities for finding approximate solutions to CVRPTW. This paper proposes a novel LLM-enhanced Q-learning framework to address the CVRPTW with real-time emergency constraints. Our solution introduces an adaptive two-phase training mechanism that transitions from the LLM-guided exploration phase to the autonomous optimization phase of Q-network. 
To ensure reliability, we design a three-tier self-correction mechanism based on the Chain-of-Thought (CoT) for LLMs: syntactic validation, semantic verification, and physical constraint enforcement. In addition, we also prioritized replay of the experience generated by LLMs to amplify the regulatory role of LLMs in the architecture. Experimental results demonstrate that our framework achieves a 7.3\% average reduction in cost compared to traditional Q-learning, with fewer training steps required for convergence.
\end{abstract}


\section{INTRODUCTION}

The Vehicle Routing Problem (VRP) is a classic problem in operations research, where single or multiple trucks must travel from a depot to satisfy customer demands and return to the depot. The objective is to design routes for the vehicles that minimize the total cost. The Capacitated Vehicle Routing Problem (CVRP) extends the traditional VRP by adding capacity constraints to the vehicles, making it more realistic. Furthermore, the Capacitated Vehicle Routing Problem with
Time Windows (CVRPTW) incorporates time constraints, where customers require service within a specific time window \cite{toth2014vehicle}. The difficulty of these problems is to find exact solutions or a high-quality approximate solutions in a short time.

\begin{figure}
  \centering
  \includegraphics[width=0.5\linewidth]{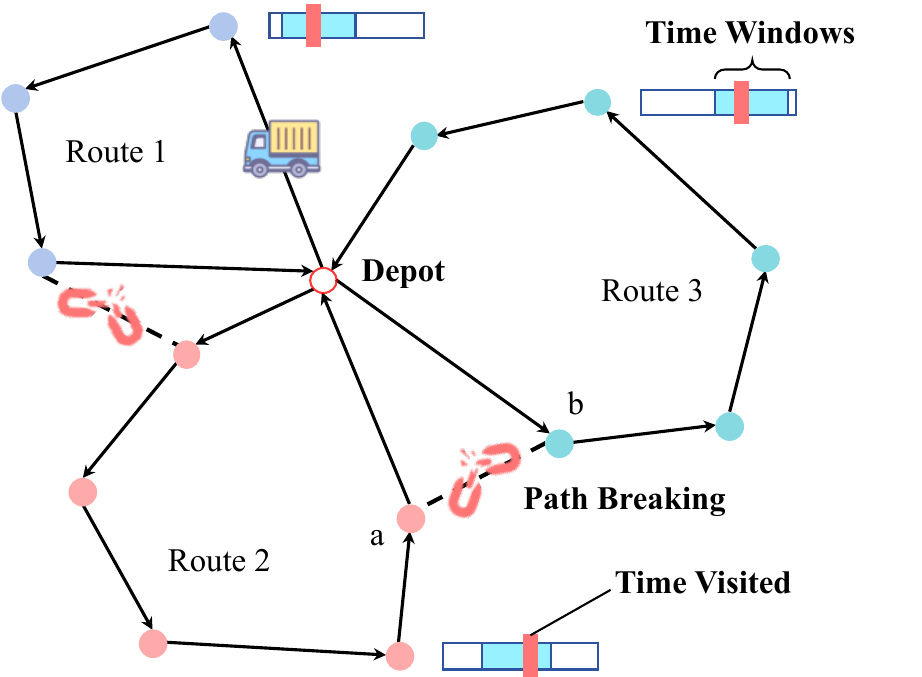}
  \caption{Schematic of CVRPTW with path breaking constraints. }
  \label{fig:CVRPTW}
\end{figure}

In addition to time constraints, this paper introduces path breaking within the CVRPTW framework to evaluate the system’s resilience under disrupted conditions.
Fig.~\ref{fig:CVRPTW} illustrates an example of a CVRPTW with path constraints, where a truck follows routes 1, 2, and 3 in sequence to meet the time windows. After passing point $a$, the truck, which could have reached point $b$, alters its route and returns to the depot due to a disruption in the path. We propose a framework combining Large Language Models (LLMs) and Reinforcement Learning (RL) to solve this problem. By fully using the natural language understanding capabilities of LLMs and the iterative learning capabilities of neural networks, the framework can achieve improvements over traditional methods.

Existing research in this field has explored many solutions. 
Traditional methods to solve this problem include exact algorithms and heuristic methods. Bettinelli et al. \cite{bettinelli2011branch} proposed a branch-and-bound method for the multi-depot VRP, which provides an exact solution for this variant of VRP. Nguyen et al. \cite{itsc3} introduced a hierarchical heuristics algorithm for CVRPTW, which significantly reduces travel time and increases vehicle utilization. 
Based on heuristic algorithms, Figliozzi \cite{FIGLIOZZI2010668} proposed an improved construction algorithm to solve VRP with soft time windows.
However, these methods often require excessive computation time and cannot adapt to scenarios with complex or real-time constraints. 
Although heuristics have achieved significant progress in recent years, RL methods have performed well in some cases, especially after Deep Q-Network (DQN) were proposed\cite{mnih2015human}. 
Kim et al. \cite{kim2021learning} applied learning collaborative policies to CVRP. Jiang et al. \cite{NEURIPS2023_a68120d2} proposed an ensemble-based RL for VRP. Their research shows the great potential of the RL approach for solving such problems. 
Research progress of LLMs provides new ideas for RL in enhanced combinatorial optimization. LLMs are good at parsing semantic constraints, generating structured action spaces, and providing domain-specific heuristics through natural language reasoning \cite{itsc2}. 
The combined algorithm of LLM and RL breaks through the limitations of traditional algorithms. Luo et al. \cite{NEURIPS2023_1c10d0c0} proposed a generalization method for CVRP based on large models and neural optimization. Alsadat et al. \cite{alsadat2024multi} applied LLMs to enhance multi-agent coordination in stochastic games. Yan et al. \cite{yan2025hybrid} designed a LLM-RL framework applied to vehicle routing and communication, and it is proved that this improvement can effectively improve the satisfaction rate of time window constraints. 
However, existing LLM-enhanced RL approaches often fail to take full advantage of the Chain of Thought (CoT) or avoid hallucinations of LLM. 

Based on these studies, we explore a new LLMs-enhanced Q-learning framework for CVRPTW. By the semantic reasoning capability of LLMs and the iterative learning capability of neural networks, this framework provides a simple and practical approach to solve CVRPTW under real-time constraints. Our framework can balance exploration and exploitation in complex action spaces without the need of fine-tuning or pre-training. The proposed method introduces a two-stage adaptive training framework, making full use of the ability of LLMs to provide excellent initial solutions and the learning ability of RL, achieving policy improvement while maintaining training stability. This study emphasizes the potential of an interactive inference optimization framework for LLMs with neural optimization methods to solve real world problems. 
The main contributions of our paper are as follows. 
\begin{itemize}
\item  A two-stage hybrid training framework is proposed, transitioning from LLM-guided exploration to autonomous optimization. LLMs analyze the spatiotemporal and capacity constraints of actions while incorporating CoT to reduce the exploration complexity of neural network. Q-learning is used to enhance the reasoning capabilities of LLMs.
\item We design a replay mechanism to prioritize LLM experience pools. The Prioritized Experience Replay (PER) module introduces LLM generation markers, increasing the probability of sampling from the LLM experience pool and amplifying the guiding role of the LLM in the training process.
\item The proposed method is validated on the classic CVRP dataset, which includes time window and path pass constraints. The results show a $7\%$ performance improvement over tradition methods, along with faster convergence.
\end{itemize}

The remainder of this paper is structured as follows. Section II formulates the problem. Section III comprehensively elaborates our proposed framework, including its core modules and processes. Section IV shows the comparison results of our method with baseline methods and analyzes the ablation study results. Finally, Section V summarizes the main conclusions of this paper and outlines possible directions for future research. 

\section{Problem Formulation} \label{02_modeling}
This section describes the problem we studied and establishes the mathematical foundation. 
\subsection{Problem Definition}
We consider three components in the CVRPTW in our paper: vehicle capacity constraint, time window constraint and path break constraint. The optimization goal is to design a vehicle dispatching plan with the minimum generalized cost while satisfying the above conditions. In this problem, we assume that time varies uniformly with the distance traveled by the vehicle, so that time can be reflected in terms of cumulative distance traveled. 

In the network $G=(V,E)$, where $V=\{0\}\cup C$ composed of depot $0$ and customer nodes $C=\{1, \ldots, N\}$, there exist bidirectional edges $E \subseteq V \times V$ between all nodes, where $E=E_{tp}\cup E_{ti}$ composed of passible edges $E_{tp}$ and impassable edges $E_{ti}$. A single vehicle completes $K$ routes from route $1$ to route $k$ in a row. The variable $T_{ij_e}$ defines the break time from node $i$ to node $j$ for route $k$. Each customer node $i \in C$ has a non-uniform demand $d_i > 0$ with time window constraints $[a_i, b_i]$ for allowable service times. 

The following constraints need to be met:
\begin{itemize}
\item vehicle load in each dispatch $\sum_{i \in P_k} d_i$ does not exceed the load limit $C_{cap}$.
\item Each customer $i \in C$ is served only once.
\item  Every demand of customers $d_i$ are met.
\item The vehicle arrived Time $T_{ik}$ at the customer point $i$ is within its time window $[a_i,b_i]$.
\item The vehicle only drives on passable edges $E_p$.
\item Each route $r_k$ must start at depot 0, visits a subset of customer nodes C, and end at depot 0.
\end{itemize}

The problem is formulated as a mixed-integer linear programming (MILP) model with the following expressions: 
\begin{align}
    \text{Minimize} \quad & \sum_{i \in V} \sum_{j \in V} \sum_{k \in K} c_{ij} x_{ijk} \label{eq:obj}, \\
    \text{subject to} \quad & \sum_{k \in K} \sum_{j \in V} x_{ijk} = 1, \quad \forall i \in C \label{eq:visit}, \\
    & \sum_{j \in V} x_{0jk} = 1, \quad \forall k \in K \label{eq:depart}, \\
    & \sum_{i \in V} x_{ihk} - \sum_{j \in V} x_{hjk} = 0, \nonumber\\
    & \quad \forall h \in C, \forall k \in K \label{eq:flow}, \\
    & \sum_{i \in V} d_i \sum_{j \in C} x_{ijk} \leq C_{\text{cap}}, \quad \forall k \in K \label{eq:cap}, \\
    & a_i \leq T_{ik} \leq b_i, \quad \forall i \in V, \forall k \in K \label{eq:time}, \\
    & T_{jk} \geq T_{ik} + t_{ij} - M(1 - x_{ijk}),\nonumber \\
    &\quad \forall i,j \in V, \forall k \in K \label{eq:time_relation},\\
    & T_{jk} - T_{ij_e} \leq Mz_{ijk}-\epsilon, \quad \forall i, j, k, \\
    & T_{ij_e} - T_{jk} \leq M(1-z_{ijk}), \quad \forall i, j, k, \\
    & x_{ijk} \leq 1-z{ijk}, \quad \forall i, j, k, \\
    & x_{ijk} \in \{0, 1\}, z_{ijk} \in \{0, 1\}, T_{ik} \geq 0, \label{eq:vars}
\end{align}
where the decision variable $x_{ijk}$ indicates whether vehicle $k$ travels from node $i$ to node $j$, $z_{ijk}$ indicates whether the edge from node $i$ to node $j$ is disrupted during the route $k$, $c_{ij}$ is the associated travel cost, $t_{ij}$ is the travel time between nodes $i$ and $j$, the large constant $M$ and small constant $\epsilon$ are used to ensure the linearity of the model.

\subsection{Markov Decision Process}
By connecting the waypoints in the path, the problem has no aftereffects when the state is determined. We can therefore model the problem as a Markov decision process \cite{puterman1990markov}: $\mathcal{M} = \langle \mathcal{S}, \mathcal{A}, \mathcal{P}, \mathcal{R}, \gamma \rangle$, where state space $\mathcal{S}$ contains $\mathcal{V}$ which captures vehicle positions and $\mathcal{D}_t$ that models stochastic demand patterns, $\mathcal{A}$ governs action space, $\mathcal{P}$ defines transition probabilities, $\mathcal{R}$ specifies reward mechanisms, and $\gamma$ is the discount factor for future rewards.

The state takes into account the vehicle location, remaining capacity, and demand satisfaction, which is defined as follows. 
\begin{equation}
s_t = (p_v, c_r, d_t) \in \mathbb{R}^2 \times [0,\mathcal C_{cap}] \times \{0,1\}^N,
\end{equation}
where $s_t$ is the state of the vehicle at time t, $p_v$ encodes vehicle coordinates in 2D operational space, $c_r$ refers to remaining cargo capacity, $d_t$ is the $N$-dimensional service status vector ($N$: total service nodes). Each element $d_t^i\in\{0,1\}$ indicates pending service at node $i$. 

The action space changes according to the vehicle states, as detailed below. 
\begin{equation}
\mathcal{A}_t = \begin{cases} 
\mathcal{N}_f \cup \{0\} & \text{if } p_v\neq p_0 \\
\mathcal{N}_a & \text{otherwise}
\end{cases},
\end{equation}
where $\mathcal{N}_f = \{i | d_i \leq c_r\}$ determines feasible deliveries based on remaining capacity $c_r$ and nodal demand $d_i$, $\mathcal{N}_a = \{i | m_t^i=1\}$ identifies active service nodes, and action 0 represents refueling operations, $p_0$ is the location of depot. 

In order to avoid the problem of sparse rewards, we give instant rewards after each step of the action is executed. Considering the inefficiency of exploration due to the large state space when the network size of the problem is large, we use reward reshaping to encourage specific exploration. Reward reshaping includes a reward for visiting new demand points to encourage exploration, a completion reward to encourage progress, and a capacity utilization reward to encourage full utilization of capacity. The multi-objective reward function comprehensively considers the path cost, vehicle scheduling cost, violation penalty and reward reshaping. 

The reward for each time step is set as follows.
\begin{equation}
\label{rewardfunction}
r_t = -\gamma_d\Delta d_t -\gamma_f\delta_{0a} - \gamma_e\psi_k(t) + r_{re},
\end{equation}
where $\gamma_d$ is distance cost coefficient, $\gamma_f$ is fixed scheduling cost coefficient, $\gamma_e$ is a big penalty coefficient for violating the time window constraint. $\Delta d_t$ represents the distance traveled at step t, $\delta_{0a}$ equals to 1 if the vehicle is departing from the depot, otherwise it is 0. the emergency penalty component $\psi_k(t)$ is activated when not within the time window $T_{ij_e}$ of the special node $\mathcal{K}_e$, $r_{re}$ represents reward reshaping, including rewards for accessing new demand points, progress proportion rewards, and capacity utilization rewards, which are much smaller than other items. Our goal is to find the optimal policy $ \pi^* $ that reaches the maximization of expected discounted cumulative reward.

\section{LLM-Enhanced Q-learning for CVRPTW}
\label{03_algorithm}
In this section, our proposed hybrid framework that combines the advantages of LLM and Q-learning is introduced in detail to solve the problem described in Fig. \ref{fig:CVRPTW}.
\subsection{Our Framework}

This hybrid architecture cleverly combines the inference capabilities of LLMs with the optimization methods of neural networks. In the early phase, by employing LLM candidate strategies as metaheuristic-inspired approaches, we can effectively enhance the exploration efficiency of neural networks in high-dimensional spaces. During the subsequent phase, we designed a novel PER mechanism to expand the influence of LLMs. In addition, our algorithm can flexibly switch stages according to environmental changes. 

\begin{figure}[htbp]
    \centering
    \includegraphics[width=0.48\textwidth]{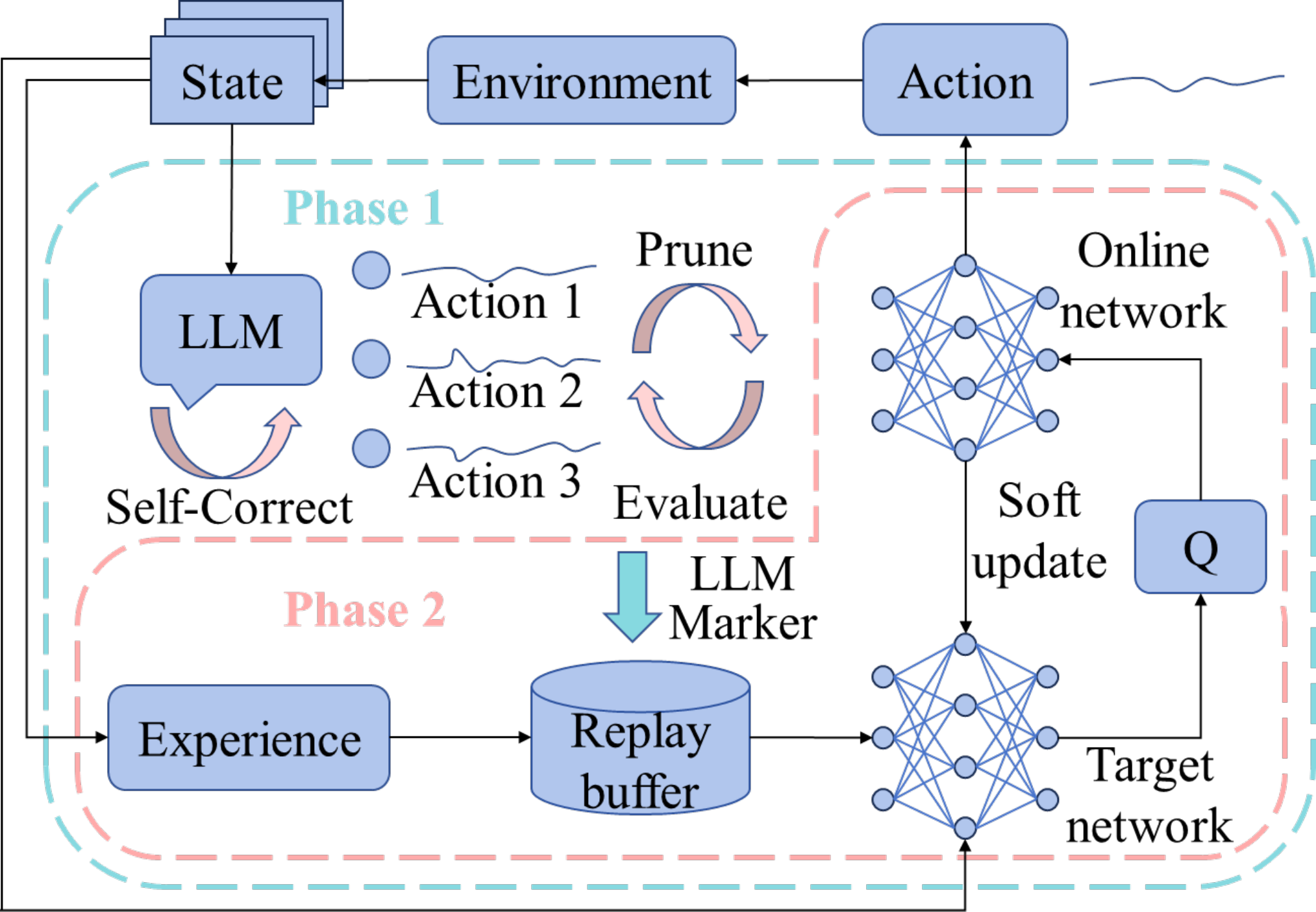} 
    \caption{LLMs-enhanced Q-learning framework.}
    \label{fig:framework}
\end{figure}

As shown in Fig. \ref{fig:framework}, our LLM-enhanced Q-learning hybrid architecture has two phases. Phase 1 is a LLM-guided exploration stage. By semantic constraint analysis, LLMs can generate three well-formed trajectory candidates $\mathcal{A}_g$, constraining the action space of DQN to $\mathcal{A}_t \cap \mathcal{A}_g$. While DQN evaluates the response of LLM through the state value and action advantage of dual neural network fitting, and finally makes decisions through online network. Here, the experience generated by the LLM stored in the replay buffer is marked. This phase is characterized by tendentious exploration guided by LLMs. Phase 2 is a autonomous refinement stage.  DQN samples freely from experience replay buffer $\mathcal{B}$ with prioritized experience replay, focuses on replaying the trajectory strategy given by the LLM. DQN dynamically adjusts the learning strategy by dynamically sensing the environment to accelerate convergence. This phase is characterized by enhanced exploitation, with more playback from previous experience guided by large models. 

\subsection{Synergistic Components}

Our LLM-enhanced Q-learning architecture addresses dynamic CVRPTW challenges with multiple synergistic components: a double DQN module, a dueling network structure for value estimation, prioritized experience replay for LLM-guided sampling, an LLM advisory module with multi-stage self-correction, and an implicit CoT LLM that maintains a memory pool. The pseudocode for
our LLM-enhanced DQN is shown in Algorithm \ref{algorithm}. We apply double DQN \cite{doubledqn} and dueling DQN \cite{duelingdqn} to our method. in double DQN method, we implement a soft update strategy for our neural networks. The parameters are updated using the polyak method. 

On the basis of the PER, we strengthen the sampling of the LLM trajectory pool by modifying the sampling priority calculation. The conversion priority $p_i$ combines temporal difference (TD) error $\delta_t$ and LLM good empirical magnification. In this way, we expect to be able to fully reflect the dominant interval of the excellent trajectory and avoid the recurrence of inefficient trajectories. The conversion priority $p_{i}$ is calculated as follows:

\begin{equation}
p_t = |\delta_t| \cdot( 1 + \epsilon_{\text{LLM}} \cdot \mathbb{I}_{\text{LLM}}) + \epsilon,
\end{equation}
where $\epsilon_{\text{LLM}}$ weights extra LLM suggestion quality, and $\mathbb{I}_{\text{LLM}}$ reflects if it is generated by LLM. The stochastic sampling probability $P(i)$ and importance-sampling weight $w_i$ used in this work are defined in accordance with the method proposed in \cite{PER}, which enables bias correction under non-uniform sampling. 


\begin{figure}[htbp]
    \centering
    \includegraphics[width=0.48\textwidth]{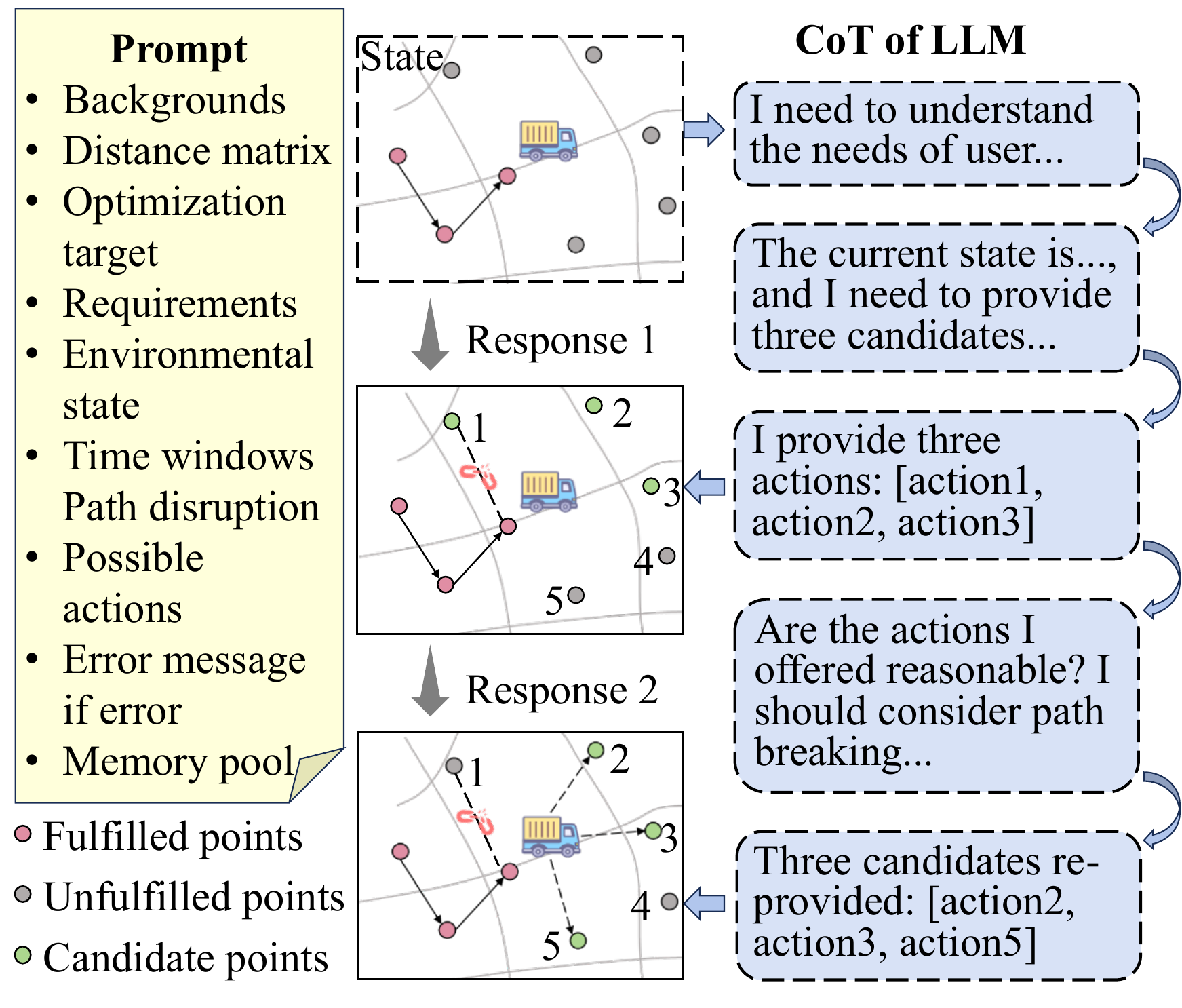} 
    \caption{Prompt and CoT of LLM in the process of generating candidates.}
    \label{fig:CoT}
\end{figure}

\begin{algorithm}[ht]
\caption{LLM-Enhanced Q-learning}
\label{algorithm}
\DontPrintSemicolon
\BlankLine
Initialize dueling network $\theta$, target network $\theta^-$, replay buffer $\mathcal{B}$, LLM memory pool $\mathcal{M}_{\text{LLM}}$ with capacity $C_{\text{m}}$\;
\For{episode = 1 \KwTo $M$}{
    Observe state $s_0$, initialize trajectory $\tau$, priority weights $\rho$\;
    \For{$t = 0$ \KwTo $H$}{
        \If{$\text{LLM\_Active}(s_t)$}{
            Candidates $\tau_g \leftarrow \text{Top-K}(\text{LLM}(s_t))$\;
            $\tau_g'\leftarrow$LLM Self-Correction\;
            $\mathcal{A}_t \leftarrow \mathcal{A}_t \cap \mathbf{\tau_g'}$\;
        }
        Select action $a_t = \epsilon\text{-greedy}(Q(s_t,\cdot;\theta))$\;
        Execute $a_t$, observe $s_{t+1}$ and shaped reward $r_t$\;
        \textbf{Priority Calculation}\;
        Calculate the TD error $\delta_t$\;
        Store $(s_t,a_t,r_t,s_{t+1})$ in $\mathcal{B}$ with priority $p_t$\;
        \If{$\text{Update\_Condition}()$}{
            Sample batch $j$ from $\mathcal{B}$\;
            Compute sampling weights $w_j$\;
            
            \textbf{Dueling Target}\;
            $Q_{\text{duel}} \leftarrow$ value function $V(s;\theta^-)$, advantage function $A(s,a;\theta^-)$\;
            
            Update $\theta$ using gradient descent with loss $\mathcal{L}$\;
            
            Update priorities $p_j$\;
            Update $\theta^-$ via Polyak averaging method
        }
    }
    Update LLM memory pool $\mathcal{M}_{\text{LLM}} \leftarrow \text{Top-K}(\tau)$\;
}
\end{algorithm}

Since the hallucination problem of LLMs can be serious, we have designed a self-correcting module to improve the reliability of LLMs in the overall architecture. The LLM self-error module has a three-stage filtering mechanism to ensure the validity of candidate trajectories by feeding back the previous reply in the memory pool with the error type. The three-layer filtering mechanism is as follows. Syntax validation layer which uses regular expression matching to filter malformed trajectories and provide feedback, semantic verification layer which detects whether the generated trajectory is reasonable and avoid hallucinations, physical Constraint Enforcement Layer which enforces environmental constraints based on spatiotemporal constraints. 

In terms of prompt engineering, we define the role of LLM as a solver and write the optimization goal into the prompt word. By synchronizing state to the LLM in real time, it is able to interact with the environment through output candidates. In addition, we strictly limit its output structure for different tasks in order to parse the content. It is worth mentioning that adding ``output in list or dictionary format'' to the prompt word will be effective. 
Fig. \ref{fig:CoT} illustrates our prompt word engineering and CoT working examples. Our prompts contain multidimensional information, error message and memory pool. By properly designing the prompts, we will be able to make the LLM implement a multi-step inference process. The combination of this CoT and three-layer error detection enables LLM self-correction. As shown in Fig. \ref{fig:CoT}, the response 1 does not satisfy the path reachability constraint, and the LLM will be able to take this reply from the memory pool to revisit it. After multiple steps of reasoning, a reasonable response 2 is given by LLM. 

\section{Numerical Results} 
\label{04_experiments}
This section provides a detailed description of our experimental setup, compared methods, performance comparison, and ablation studies. 

\subsection{Experimental Setup}
The dataset used in this experiment is Vehicle Routing Data Set A and Set B \cite{augerat1995computational}. Set A and Set B contain 27 and 23 CVRP instances of different network nodes. We add custom Time Window Constraints and Path Break Constraints on each of these problems. To improve reliability of the experiments, we performed three parallel experiments in the same configuration and averaged them for each instance. 

Through joint optimization of grid search and genetic algorithm, the configuration parameters were determined: a discount factor $\gamma$ of 0.99, a $\epsilon$ attenuation rate of 0.995, an initial $\epsilon$ of 0.8, a minimum $\epsilon$ of 0.01, and a learning rate of 0.0002. Additional setting for the LLM enhancement method: an extra LLM experience replay weight $\epsilon_{\text{LLM}}$ of 1.5. In the reward function, we take the value of $\gamma_d$ as 4.5 and the value of  $\gamma_f$ as 65, which are obtained through actual freight data. The LLaMA model we used is meta-llama/Llama-3.1-70B-Instruct and the GPT model we used is GPT-4o.
\subsection{Compared Methods}

In this paper, we compare our method with the DQN baseline method. The accuracy and constraint compliance rate of DQN, LLaMA-enhanced DQN and GPT-enhanced DQN were compared, indicating the role of LLMs in the framework. These methods are improved as follows:
Double DQN is used to decouple action selection and value evaluation to alleviate the problem of overestimation of Q value, Dueling DQN is used to separate state value and advantage functions to improve policy expression, prioritized Experience Replay is used to prioritize the replay of important samples, in which the LLM enhancement method plays back the policy focus given by the LLM. 
\subsection{Performance Comparison}

\begin{figure}[htbp]
    \centering
    \includegraphics[width=0.485\textwidth]{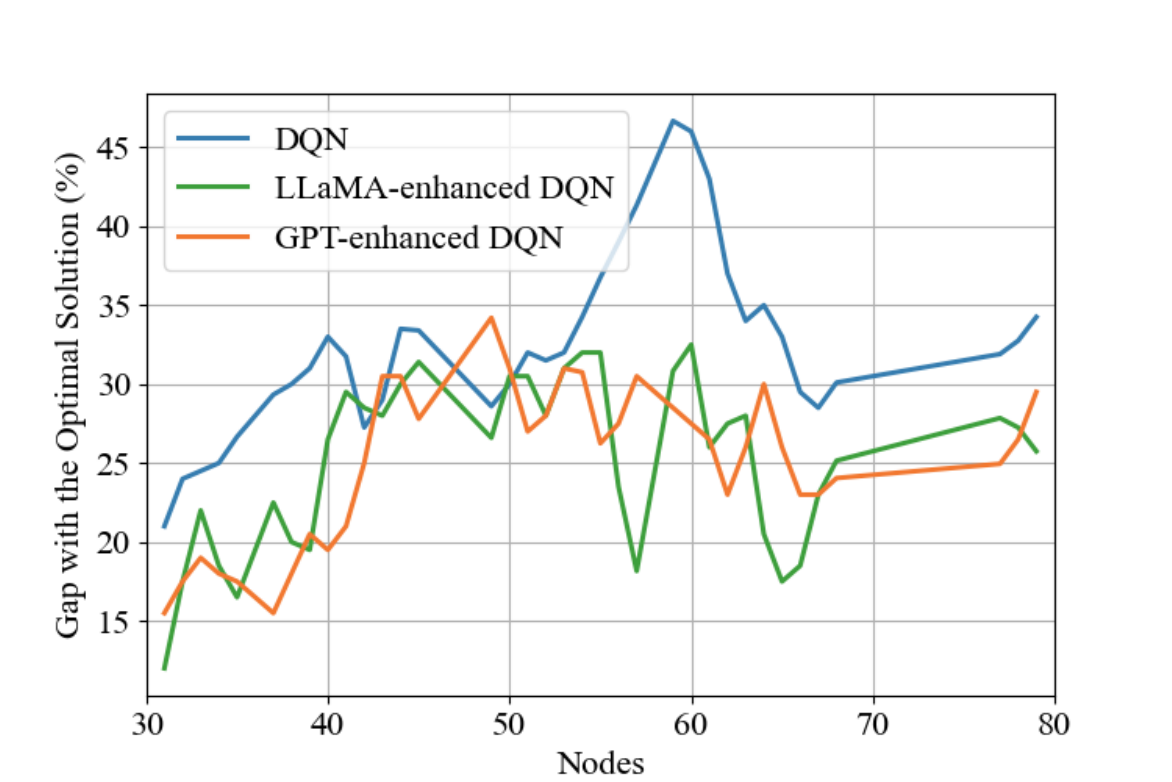} 
    \caption{Comparison of algorithm performance on different instances. Gap with the optimal solution is the ratio of the generalized cost to the optimal solution generalized cost. For parallel experiments, we take the average the results. }
    \label{fig:Variation with Network size}
\end{figure}

Fig. \ref{fig:Variation with Network size} illustrates the performance of our methods and traditional approaches. On instances of the dataset, the LLM-enhanced method achieves a better approximation in most cases. Compared to traditional DQN, our method achieves an average cost reduction of 7.3\%, with the LLaMA method decreasing by 7.0\% and the GPT method decreasing by 7.5\%. In addition, the training of our method is more stable and converges faster. However, the performance will change as the number of nodes increases. 

The variation curve of the effect of the algorithm on the dataset with the number of network nodes is also shown in Fig. \ref{fig:Variation with Network size}. It can be found that the LLMs enhancement method achieves better results on most instances, but the effect is close to that of DQN when the number of nodes is medium. This may be due to the fact that when the number of nodes is small, LLMs can infer the optimal trajectory without other algorithms, and this advantage gradually decays as the number of nodes increases. As the network scale continues, the shortcomings of DQN exploration efficiency are exposed and our method achieves better solutions.
In the instances we tested, the GPT method performed more consistently, while the LLaMA method performed quite differently in instances with different nodes.
\begin{figure}[htbp]
    \centering
    \includegraphics[width=0.485\textwidth]{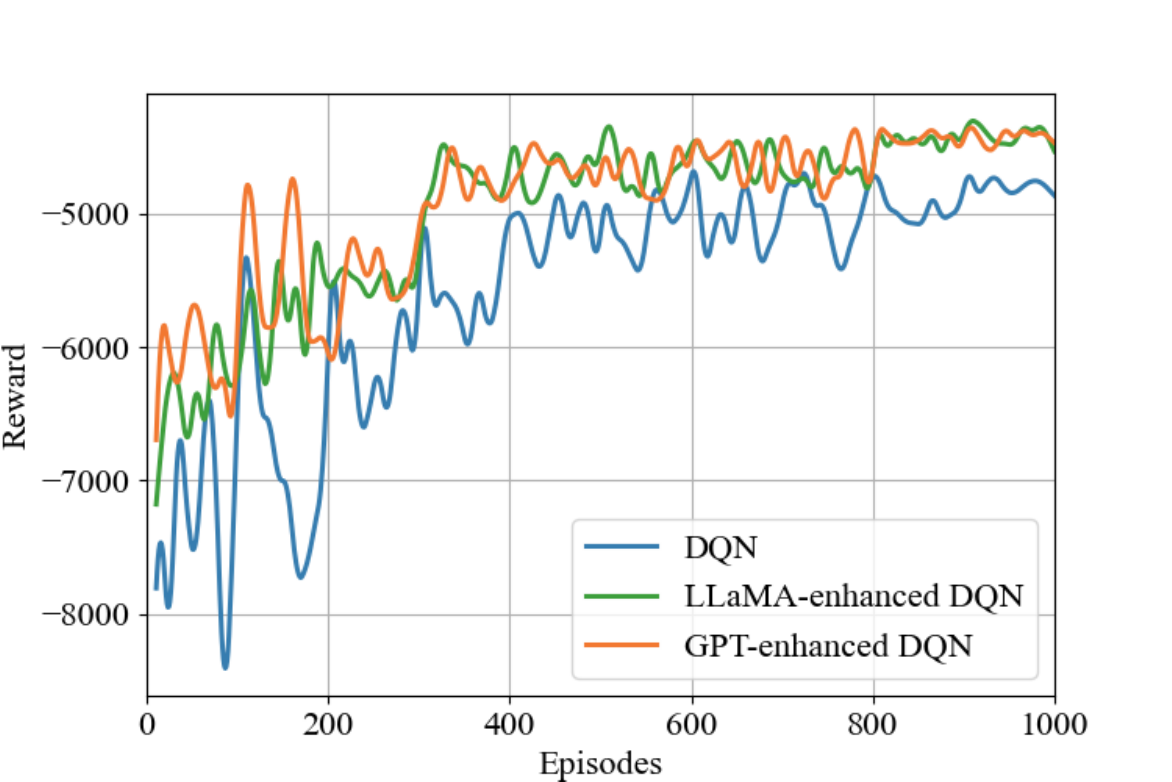} 
    \caption{Performance curves of algorithms on instance B-n34-k5.}
    \label{fig:Reward Curse}
\end{figure}

Fig. \ref{fig:Reward Curse} is a training process reward curve on an example that shows that the LLMs-enhanced DQN approach works better in the initial solution and converges faster. In terms of the quality of the approximate solution finally explored, the LLM-enhanced DQN method has a lower generalized cost and is better than the traditional DQN method. In this instance, GPT-enhanced DQN explored a better solution in the early stage, while both LLM variants achieve comparable final costs.

\subsection{Ablation Study}

Table~\ref{tab:accuracy_comparison} and Table~\ref{tab:constraint_satisfaction_rate_comparison} present the results of the ablation study, which examines the impact of each module, including LLM memory, prioritized experience replay (PER), double DQN, dueling DQN and the reward reshape module. 

The removal of the LLM memory module led to a $5.6\%$ increase in the cost of the LLaMA-enhanced DQN and a $4.8\%$ increase in the cost of the GPT-enhanced DQN, highlighting the importance of the memory pool. Additionally, the removal of the PER module resulted in a significant increase in the average gap and a noticeable decrease in constraint compliance for the LLM-enhanced methods compared to the benchmark DQN, indicating that the high-quality experiences generated by the LLM are indispensable for handling complex constraints. The removal of double DQN caused a significant performance decline across all methods, confirming its essential role. When the dueling network is ablated, the average gap of LLaMA-enhanced DQN increased by $2.6\%$, while GPT-enhanced DQN accuracy increased by $4.0\%$, which seems to mean it is not that important. Similarly, removing the reward reshape module led to a slight decrease in the performance of all methods, demonstrating that the value-state separation in the dualing network enhances action value estimation.. 

We can conclude that the LLM memory pool, LLM PER, and double DQN modules play a key role in our method, while the dueling DQN and reward reshape modules are less significant. 

\begin{table}[htbp]
\centering
\caption{Average Gap$^{\mathrm{a}}$\tnote{a} Comparison (\%)}
\label{tab:accuracy_comparison}
\begin{tabular}{lcccc}
\toprule
\textbf{Configuration} & \textbf{DQN} & \textbf{LLaMA-DQN} & \textbf{GPT-DQN} \\
\cmidrule{1-4}
All & 31.9 & 24.9 & 24.4 \\
No LLM memory & \multicolumn{1}{c}{—} & 30.5 & 29.2 \\
No LLM PER & \multicolumn{1}{c}{—} & 28.8 & 29.7 \\
No double DQN & 44.7 & 38.6 & 41.1 \\
No dueling DQN & 34.6 & 27.5 & 28.4 \\
No reward reshape & 32.4 & 26.9 & 26.3 \\
\bottomrule
\end{tabular}

\parbox{\textwidth}{
\vspace{2mm}
\hspace*{140pt}
\begin{tabular}{@{}p{\textwidth}@{}}
{$^\mathrm{a}$} \textit{Average gap} is computed based on the best solutions. 
\end{tabular}}

\end{table}

\begin{table}[htbp]
\centering
\caption{Constraint Satisfaction Rate Comparison (\%)}
\label{tab:constraint_satisfaction_rate_comparison}
\begin{tabular}{lcccc}
\toprule
\textbf{Configuration} & \textbf{DQN} & \textbf{LLaMA-DQN} & \textbf{GPT-DQN} \\
\cmidrule{1-4}
All & 63.3 & 81.3 & 82.7 \\
No LLM memory & \multicolumn{1}{c}{—} & 51.3 & 50.3 \\
No LLM PER & \multicolumn{1}{c}{—} & 46.7 & 49.3 \\
No double DQN & 29.3 & 39.3 & 42.7 \\
No dueling DQN & 55.7 & 78.3 & 75.7 \\
No reward reshape & 63.7 & 77.3 & 80.3 \\
\bottomrule
\end{tabular}
\smallskip
\footnotesize
\end{table}

\section{CONCLUSIONS}
\label{05_conclusions}
In this work, we explore the capability boundaries of LLMs in classical NP-hard combinatorial optimization problems. We improve the inference capabilities of LLMs through a novel LLM-RL architecture. By combining the semantic reasoning ability of LLM with the RL framework, this paper improves the accuracy of CVRPTW and greatly improves the compliance rate of constraints. The ablation experiments show that the LLM memory module helps to improve the performance, and the reward remodeling plays a small role in the algorithm improvement.

This work can be extended by incorporating additional constraints, such as a heterogeneous fleet. We also plan to investigate the integration of large language models (LLMs) with combinatorial optimization techniques. Furthermore, future research will explore the use of graph neural networks (GNNs) to capture spatial correlations, which we believe represents a promising direction for advancing this field.

\bibliographystyle{ieeetr}
\bibliography{main}

\begin{thebibliography}{10}

\bibitem{toth2014vehicle}
P.~Toth and D.~Vigo, {\em Vehicle routing: problems, methods, and applications}.
\newblock SIAM, 2014.

\bibitem{bettinelli2011branch}
A.~Bettinelli, A.~Ceselli, and G.~Righini, ``A branch-and-cut-and-price algorithm for the multi-depot heterogeneous vehicle routing problem with time windows,'' {\em Transportation Research Part C: Emerging Technologies}, vol.~19, no.~5, pp.~723--740, 2011.

\bibitem{itsc3}
Q.~C. Nguyen, L.~Yang, Z.~Li, and W.~Chen, ``Hierarchical vehicle routing for online delivery platform,'' in {\em 2019 IEEE Intelligent Transportation Systems Conference (ITSC)}, pp.~1709--1714, 2019.

\bibitem{FIGLIOZZI2010668}
M.~A. Figliozzi, ``An iterative route construction and improvement algorithm for the vehicle routing problem with soft time windows,'' {\em Transportation Research Part C: Emerging Technologies}, vol.~18, no.~5, pp.~668--679, 2010.

\bibitem{mnih2015human}
V.~Mnih, K.~Kavukcuoglu, D.~Silver, A.~A. Rusu, J.~Veness, M.~G. Bellemare, A.~Graves, M.~Riedmiller, A.~K. Fidjeland, G.~Ostrovski, {\em et~al.}, ``Human-level control through deep reinforcement learning,'' {\em nature}, vol.~518, no.~7540, pp.~529--533, 2015.

\bibitem{kim2021learning}
M.~Kim, J.~Park, and j.~kim, ``Learning collaborative policies to solve np-hard routing problems,'' in {\em Advances in Neural Information Processing Systems}, vol.~34, pp.~10418--10430, 2021.

\bibitem{NEURIPS2023_a68120d2}
Y.~JIANG, Z.~Cao, Y.~Wu, W.~Song, and J.~Zhang, ``Ensemble-based deep reinforcement learning for vehicle routing problems under distribution shift,'' in {\em Advances in Neural Information Processing Systems}, vol.~36, pp.~53112--53125, 2023.

\bibitem{itsc2}
X.~Han, Q.~Yang, X.~Chen, X.~Chu, and M.~Zhu, ``Generating and evolving reward functions for highway driving with large language models,'' in {\em 2024 IEEE 27th International Conference on Intelligent Transportation Systems (ITSC)}, pp.~831--836, 2024.

\bibitem{NEURIPS2023_1c10d0c0}
F.~Luo, X.~Lin, F.~Liu, Q.~Zhang, and Z.~Wang, ``Neural combinatorial optimization with heavy decoder: Toward large scale generalization,'' in {\em Advances in Neural Information Processing Systems}, vol.~36, pp.~8845--8864, 2023.

\bibitem{alsadat2024multi}
S.~{Meshkat Alsadat} and Z.~Xu, ``Multi-agent reinforcement learning in non-cooperative stochastic games using large language models,'' {\em IEEE Control Systems Letters}, vol.~8, pp.~2757--2762, 2024.

\bibitem{yan2025hybrid}
Z.~Yan, H.~Zhou, H.~Tabassum, and X.~Liu, ``Hybrid llm-ddqn-based joint optimization of v2i communication and autonomous driving,'' {\em IEEE Wireless Communications Letters}, vol.~14, no.~4, pp.~1214--1218, 2025.

\bibitem{puterman1990markov}
M.~L. Puterman, ``Markov decision processes,'' {\em Handbooks in operations research and management science}, vol.~2, pp.~331--434, 1990.

\bibitem{doubledqn}
H.~v. Hasselt, A.~Guez, and D.~Silver, ``Deep reinforcement learning with double q-learning,'' in {\em Proceedings of the Thirtieth AAAI Conference on Artificial Intelligence}, pp.~2094--2100, 2016.

\bibitem{duelingdqn}
Z.~Wang, T.~Schaul, M.~Hessel, H.~Hasselt, M.~Lanctot, and N.~Freitas, ``Dueling network architectures for deep reinforcement learning,'' in {\em International conference on machine learning}, pp.~1995--2003, 2016.

\bibitem{PER}
T.~Schaul, J.~Quan, I.~Antonoglou, and D.~Silver, ``Prioritized experience replay,'' in {\em 4th International Conference on Learning Representations, {ICLR} 2016}, 2016.

\bibitem{augerat1995computational}
P.~Augerat, D.~Naddef, J.~Belenguer, E.~Benavent, A.~Corberán, and G.~Rinaldi, ``Computational results with a branch and cut code for the capacitated vehicle routing problem,'' Tech. Rep. INPG-RR-949-M, Institut National Polytechnique, 38 - Grenoble, 1995.

\end{thebibliography}
\end{document}